\documentclass{article}
\usepackage{spconf}
\usepackage{times}
\usepackage{epsfig}
\usepackage{graphicx}
\usepackage{amsmath}
\usepackage{amsthm}
\usepackage{amssymb}
\usepackage{amsfonts}
\usepackage{url}
\usepackage{algorithm}
\usepackage{algorithmicx}
\usepackage{algpseudocode}
\usepackage{booktabs}
\usepackage{threeparttable}
\usepackage{multirow}
\usepackage{subcaption}
\usepackage{array}
\usepackage{color}
\usepackage{soul}
\usepackage{footnote}
\usepackage{tablefootnote}
\usepackage{graphicx}

\makeatletter
\newcommand*\bigcdot{\mathpalette\bigcdot@{.5}}
\newcommand*\bigcdot@[2]{\mathbin{\vcenter{\hbox{\scalebox{#2}{$\m@th#1\bullet$}}}}}
\makeatother

\title{Multi-head Uncertainty Inference for Adversarial Attack Detection}
\name{Yuqi Yang$^{1}$, Songyun Yang$^{1}$, Jiyang Xie$^{1}$, Zhongwei Si$^{1,*}$, Kai Guo$^{1}$, Ke Zhang$^{2,3}$, Kongming Liang$^{1}$\thanks{*: Corresponding author: sizhongwei@bupt.edu.cn
}}
\address{$^{1}$Beijing University of Posts and Telecommunications, Beijing, China \\$^{2}$Department of Electronic and Communication Engineering,\\North China Electric Power University, Baoding, Hebei, China \\$^{3}$Hebei Key Laboratory of Power Internet of Things Technology,\\North China Electric Power University, Baoding, Hebei, China}

\begin{document}

\topmargin=0mm

\maketitle%

\begin{abstract}
Deep neural networks (DNNs) are sensitive and susceptible to tiny perturbation by adversarial attacks which causes erroneous predictions. Various methods, including adversarial defense and uncertainty inference (UI), have been developed in recent years to overcome the adversarial attacks. In this paper, we propose a multi-head uncertainty inference (MH-UI) framework for detecting adversarial attack examples. We adopt a multi-head architecture with multiple prediction heads (\emph{i.e.}, classifiers) to obtain predictions from different depths in the DNNs and introduce shallow information for the UI. Using independent heads at different depths, the normalized predictions are assumed to follow the same Dirichlet distribution, and we estimate distribution parameter of it by moment matching. Cognitive uncertainty brought by the adversarial attacks will be reflected and amplified on the distribution. Experimental results show that the proposed MH-UI framework can outperform all the referred UI methods in the adversarial attack detection task with different settings.
\end{abstract}
\section{Introduction}
\label{sec:intro}

Deep neural networks (DNNs) are powerful learning tools in computer vision such as image recognition~\cite{gu2015recent} and image segmentation~\cite{badrinarayanan2017segnet}. However, DNNs have a inevitable problem which concerns with the their stability with respect to tiny perturbations on their input images that may be perceived by human eyes~\cite{elsayed2018adversarial}. Their predictions may be arbitrarily affected by the random or targeted imperceptible perturbation.

Adversarial attack task~\cite{szegedy2013intriguing} was proposed to study the best way on confusing the DNNs by generating adversarial attack examples with the aforementioned perturbations,~\emph{e.g.}, fast gradient sign method (FGSM)~\cite{goodfellow2014explaining}. These perturbations are commonly found by optimizing the input images to maximize the prediction errors and minimize additional perturbations simultaneously~\cite{ren2020adversarial},~\emph{i.e.}, white-box attacks. The adversarial attack examples can greatly affect the security and reliability of the DNNs. 

In recent years, in order to overcome the adversarial attacks, adversarial defense methods usually concentrate on random smoothing~\cite{salman2020denoised}, adversarial training~\cite{bai2021recent}, and Jacobian regularization~\cite{Jakubovitz2018improving}. Other work~\cite{xie2021dsui,gal2015dropout,malinin2018predictive,malinin2019reverse} introduced uncertainty inference (UI) for misclassification and out-of-domain detection and estimated the prediction uncertainty of a DNN, which can be treated as defense techniques for the adversarial attacks. Compared with the original adversarial defense methods that were proposed for one specific adversarial attack method, the UI methods can be commonly adapted to dispose different adversarial attack methods ~\cite{lecuyer2018certified,raghunathan2018certified}.

Most of the recently proposed UI methods can be divided into Monte-Carlo (MC) sampling based methods~\cite{gal2015dropout} and direct estimation based ones~\cite{xie2021dsui,malinin2018predictive,malinin2019reverse}. The former utilized dropout~\cite{hinton2012improving} to generate multiple subnetworks and obtained samples of predictions by them to compute variance as the uncertainty, which is computational expensive. The latter enhanced the training procedure by additional loss functions~\cite{xie2021dsui,malinin2018predictive,malinin2019reverse} and/or modules~\cite{xie2021dsui}, and directly inferred the uncertainty with the one-shot predictions, which is highly dependent on the learning ability of a DNN itself~\cite{xie2021dsui} and may be affected by the quality of extracted features. Therefore, how to construct a UI model that overcomes both of the drawbacks should be carefully investigated.

\begin{figure*}[htb]
    \begin{center}
        \includegraphics[width=.85\linewidth]{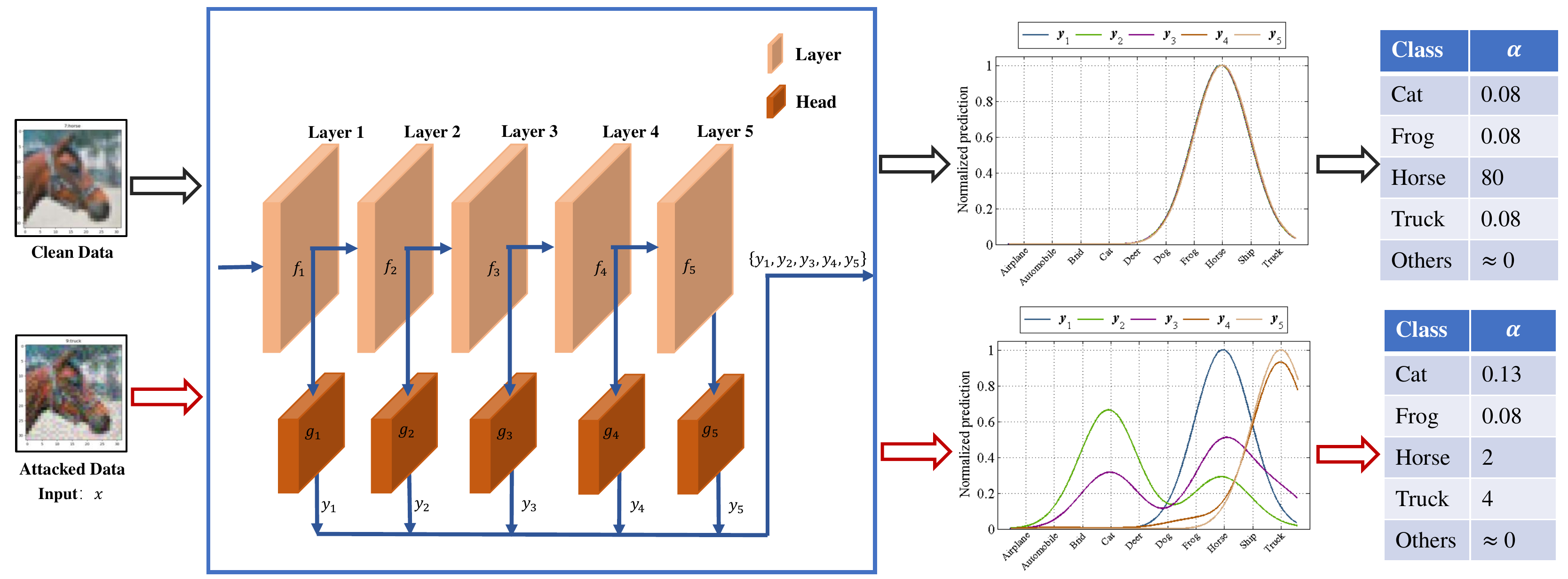}
    \end{center}
    \caption{Structure of the proposed multi-head uncertainty inference (MH-UI). We take a DNN with $5$ layers for image recognition as an example. Images are input into the multi-head architecture-based model and obtain multiple predictions from each head. For the clean data input, they usually gather in one specific class, while those of the adversarial data respectively focus on distinct classes. The predictions are treat as samples for Dirichlet distribution estimation.}
    \label{fig:structure}
\end{figure*}

On the basis of the previous work~\cite{xie2021dsui,gal2015dropout,malinin2018predictive,malinin2019reverse}, we propose to perform uncertainty analysis on the output of the DNNs to determine whether the corresponding input image is an adversarial attack example. In this paper, we propose a Multi-head Uncertainty Inference (MH-UI) framework for detecting the adversarial attack examples. As the shallower layers extract texture information and the adversarial noises usually reflect in the texture, we adapt a multi-head architecture, which is similar to the GoogLeNet-style one with multiple prediction heads~\cite{szegedy2014going}, to introduce shallow information and obtain predictions from different depths in a DNN. Independent classifiers (\emph{i.e.}, fully connected (FC) layers) are utilized in the different depths and the predictions of each are assumed to be identically Dirichlet distributed samples where distribution parameters are estimated by moment matching. Experimental results show that the proposed MH-UI framework can outperform all the referred UI methods in the adversarial attack detection task with different settings.

\section{Multi-head Uncertainty Inference}
\label{sec:mhui}

\subsection{Multi-head Architecture}
\label{ssec:multi-head}

In this section, we introduce the multi-head architecture for the MH-UI framework. Figure1 shows that a DNN stacked by $N$ layers (blocks) for image recognition can be represented as 
\begin{align}
    \boldsymbol{y}_N&=g_N \odot f(\boldsymbol{x})\nonumber\\
    &=g_N \odot f_N \odot f_{N-1} \odot \cdots \odot f_2 \odot f_1(\boldsymbol{x}),
\end{align}
where $\boldsymbol{x}\in R^{3 \times W \times H}$ ($W$ and $H$ are width and height, respectively) and $\boldsymbol{y}_N\in R^{C}$ ($C$ is class number) are an input image and the corresponding normalized predictions, respectively, $f_n,n=1,\cdots,N$ is the $n^{th}$ layer, and $\odot$ is the stack operation. And $g_N$ is the classifier of the DNN, which is constructed by multiple FC layers with a Softmax layer behind.

Recalled that the shallower layers extract texture information and the adversarial noises usually reflect in the texture. The texture information in the shallower layers should be helpful for resisting the adversarial noises and detecting the adversarial attack samples. Thus, we introduce auxiliary classifier, namely head, for each layer in order to access information from different semantic levels. Here, we define their corresponding normalized predictions $\left\{\boldsymbol{y}_1,\boldsymbol{y}_2,\cdots,\boldsymbol{y}_{N-1}\right\}$ ($\boldsymbol{y}_n\in R^{C},n=1,\cdots,N$) as
\begin{align}
    \boldsymbol{y}_n=g_n \odot f_n \odot f_{n-1} \odot \cdots \odot f_2 \odot f_1(\boldsymbol{x}),
\end{align}
where $g_n$ is the head for the $n^{th}$ layer. Note that $f_n$ can be set as any architecture, not only inception block~\cite{szegedy2014going}.

Since $\boldsymbol{y}_n$ is normalized, we roughly assume that it follows Dirichlet distribution, which is a common assumption~\cite{malinin2018predictive,malinin2019reverse}. Here, we further define all the normalized predictions of the heads ($g_N$ can be also considered as a head) as samples generated from an identical Dirichlet distribution $Dir(\boldsymbol{\alpha})$ with $C$-dimensional distribution parameter vector $\boldsymbol{\alpha}$. By further analyzing all the predictions, we can obtain the uncertainty of $\boldsymbol{x}$ and apply the uncertainty for final adversarial attack detection. 

\subsection{Uncertainty Inference}
\label{ssec:ui}

In this section, we introduce the way to infer the uncertainty of the input image $\boldsymbol{x}$. After training the whole model $f$ with the heads $\{g_1,\cdots,g_N\}$, the identically Dirichlet distributed predictions $\boldsymbol{y}=\left\{\boldsymbol{y}_1, \boldsymbol{y}_2,\cdots,\boldsymbol{y}_{N-1}\right\}$ are obtained. As defining $Dir(\boldsymbol{\alpha})$ for $\boldsymbol{y}_n$, we then estimate the parameter $\boldsymbol{\alpha}=\left\{\alpha_1,\alpha_2,\cdots,\alpha_C\right\}$ by the aforementioned samples. 

\begin{table*}
\caption{\centering{Means of AUROC (\%) on CIFAR-$10$ dataset of different head combinations under FGSM attack. The leftmost column of the table represents the combined sequence of the different heads as $1$: head $\left\{1-3,10\right\}$; $2$: head $\left\{7-10\right\}$; $3$: head $\left\{4-10\right\}$; $4$: head $\left\{1-6,10\right\}$; $5$: head $\left\{1-10\right\}$. The numbers in the brackets represent the heads involved in the uncertainty calculation. Head $1$ is the head for the first inception (closest to input). Head $10$ is the final output of the model.}  
}
\centering
\begin{tabular}{c|cccccccccc} 
\hline
\multirow{2}{*}{} & \multicolumn{2}{c}{$\boldsymbol{\epsilon} = 0.05$} & \multicolumn{2}{c}{$\boldsymbol{\epsilon} = 0.1$} & \multicolumn{2}{c}{$\boldsymbol{\epsilon} = 0.25$} & \multicolumn{2}{c}{$\boldsymbol{\epsilon} = 0.5$} & \multicolumn{2}{c}{$\boldsymbol{\epsilon} = 1.0$}  \\
                  & Max.P.~ & Ent.                                     & Max.P.~ & Ent.                                    & Max.P.~ & Ent.                                     & Max.P.~ & Ent.                                    & Max.P.~ & Ent.                                     \\ 
\hline
$1$               & 69.03   & 69.77                                    & 74.29   & 75.29                                   & 83.17   & 84.36                                    & 91.13   & 90.80                                   & 89.35   & 91.29                                    \\
$2$               & 77.52   & 77.56                                    & 79.91   & 80.02                                   & 84.06   & 84.20                                    & 85.36   & 85.41                                   & 95.69   & 96.00                                    \\
$3$               & 76.98   & 77.06                                    & 80.83   & 81.19                                   & 86.66   & 87.04                                    & 92.35   & 91.89                                   & 97.90   & 97.78                                    \\
$4$               & 71.12   & 71.48                                    & 77.79   & 78.47                                   & 87.33   & 88.33                                    & 94.47   & 95.33                                   & 97.61   & 98.44                                    \\
$5$               & 82.70   & 80.55                                    & 83.52   & 83.93                                   & 87.90   & 88.91                                    & 94.31   & 95.32                                   & 99.04   & 99.19                                    \\
\hline
\end{tabular}
\label{table3}
\end{table*}

\begin{table*} [!t]
\centering
\caption{Means and standard deviations of accuracy (\%) of heads in MH-UI  on CIFAR-$10$ and SVHN datasets.}
\resizebox{\linewidth}{!}{
\begin{tabular}{c | c c c c c c c c c} 
\hline
Head & Head$1$ & Head$2$ & Head$3$ & Head$4$ & Head$5$ & Head$6$ & Head$7$ & Head$8$ & Head$9$ \\
\hline
CIFAR-$10$ & $67.6\pm1.1$ & $71.6\pm1.3$ & $81.6\pm1.1$ & $84.4\pm1.2$ & $88.3\pm 0.8$ & $89.6\pm0.5$ & $92.0\pm 0.4$ & $93.5 \pm 0.4$ & $93.2\pm 0.6$\\
SHVN & $65.8\pm3.0$ & $69.6\pm2.7$ & $79.8\pm2.3$ & $85.1\pm 1.5$ & $88.6\pm1.2$ & $91.3\pm 1.1$ & $92.5\pm0.7$ & $94.6\pm0.3$ & $95.2\pm0.3$\\
\hline
\end{tabular}}
\label{table1}
\end{table*}

Here, we match the first- and second-order moments between $\boldsymbol{y}$ and $Dir(\boldsymbol{\alpha})$ according to the moment matching method in~\cite{xie2018balson} by
\begin{align}\label{eq:moment_matching}
\begin{cases}
    \text{E}\left[\boldsymbol{y}\right]=\frac{\boldsymbol{\alpha}}{\alpha_0}\\
    \text{Var}\left[\boldsymbol{y}\right] =\frac{\boldsymbol{\alpha}\bigcdot(\alpha_0-\boldsymbol{\alpha})}{\alpha_0^2(\alpha_0+1)}
\end{cases},
\end{align}
where $\alpha_0=\sum_{n=1}^N\alpha_n$ and $\bigcdot$ is Hadamard product. The expectation and variance of $\boldsymbol{y}$ are ddenoted by $\text{E}\left[\boldsymbol{y}\right]$ and $\text{Var}\left[\boldsymbol{y}\right]$, respectively, and are defined as
\begin{align}
\begin{cases}
    \text{E}\left[\boldsymbol{y}\right]=\frac{1}{N}\sum_{n=1}^N\boldsymbol{y}_n\\
    \text{Var}\left[\boldsymbol{y}\right]=\frac{1}{N-1}\sum_{n=1}^N(\boldsymbol{y}_n-\text{E}\left[\boldsymbol{y}\right])^2
\end{cases}.
\end{align}
According to~\eqref{eq:moment_matching}, we can obtain the value of $\boldsymbol{\alpha}$. The uncertainty of an input image can be evaluated by max probability (Max.P.$=\max_{n}m_n$) and entropy (Ent.$=-\sum_{n=1}^{N}m_n\ln(m_n)$), and $m_n=\frac{\alpha_n-1}{\alpha_0-N}$. When Max.P is larger or Ent. is smaller than the threshold value, we consider the image is attacked.

The general adversarial attack method is to confuse the final predictions, which accumulate errors from shallower layers and work with those of other heads difficultly. Such attack samples have large cognitive uncertainty. Since each head represents different depth of the DNN, their features are distinct. Therefore, some heads will become less confident when encountering the attack samples and the possibility of error recognition of the final predictions will be increased. When taking into account them of other heads, which performs similarly to an ensemble model, the aggregated predictions can be corrected. The change brought by the increase of cognitive uncertainty to the head output can be reflected on $\boldsymbol{\alpha}$ and then amplified, which allows us to evaluate $\boldsymbol{\alpha}$ to determine whether the input image is attacked.

\begin{table*} [!t]
\centering
\caption{Means and standard deviations of accuracy (\%) of each method on CIFAR-$10$ and SVHN datasets.}
\begin{tabular}{c | c c c c c c} 
\hline
Method & MC dropout & DPN & RKL & DS-UI & MH-UI (ours) \\ 
\hline
CIFAR-$10$ & $93.22\pm0.67$ & $93.55\pm0.34$ & $93.67\pm0.58$ & $92.89\pm 0.77$ & $93.11\pm0.61$ \\
SVHN & $93.60\pm0.14$ & $93.29\pm0.23$ & $93.32\pm0.24$ & $93.50\pm 0.34$ & $94.12\pm0.11$ \\
\hline
\end{tabular}
\label{table2}
\end{table*}

\begin{figure*}[!t]
    \centering
    \resizebox{\linewidth}{!}{
    \begin{tabular}{l}
    \includegraphics[width=\linewidth]{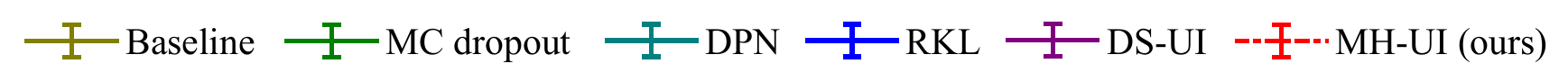} \\
    \begin{subfigure}[t]{0.49\linewidth}
        \centering
        \includegraphics[width=1\linewidth]{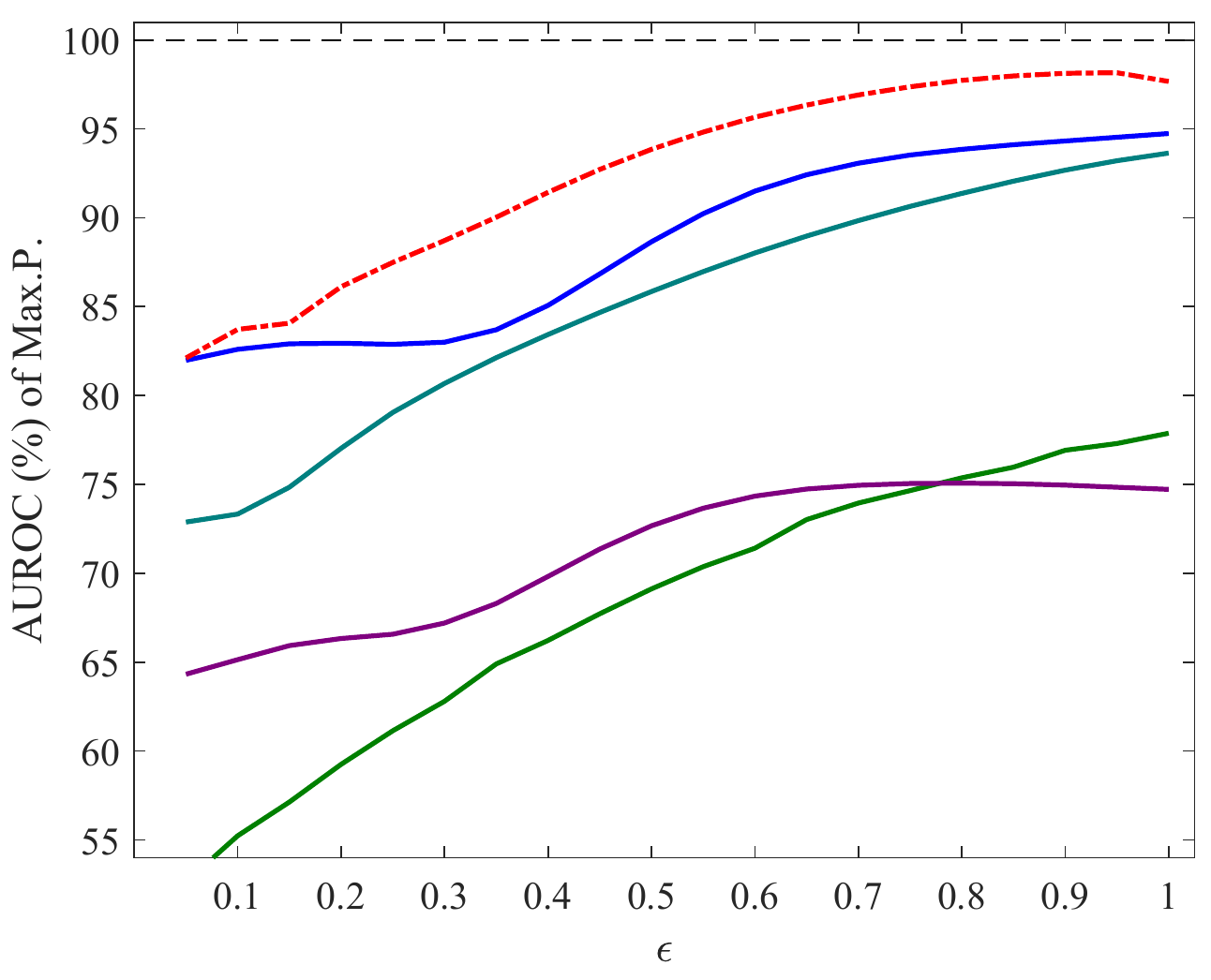}
        \subcaption{Max.P. on CIFAR-$10$ }
    \end{subfigure}
    \begin{subfigure}[t]{0.49\linewidth}
        \centering
        \includegraphics[width=1\linewidth]{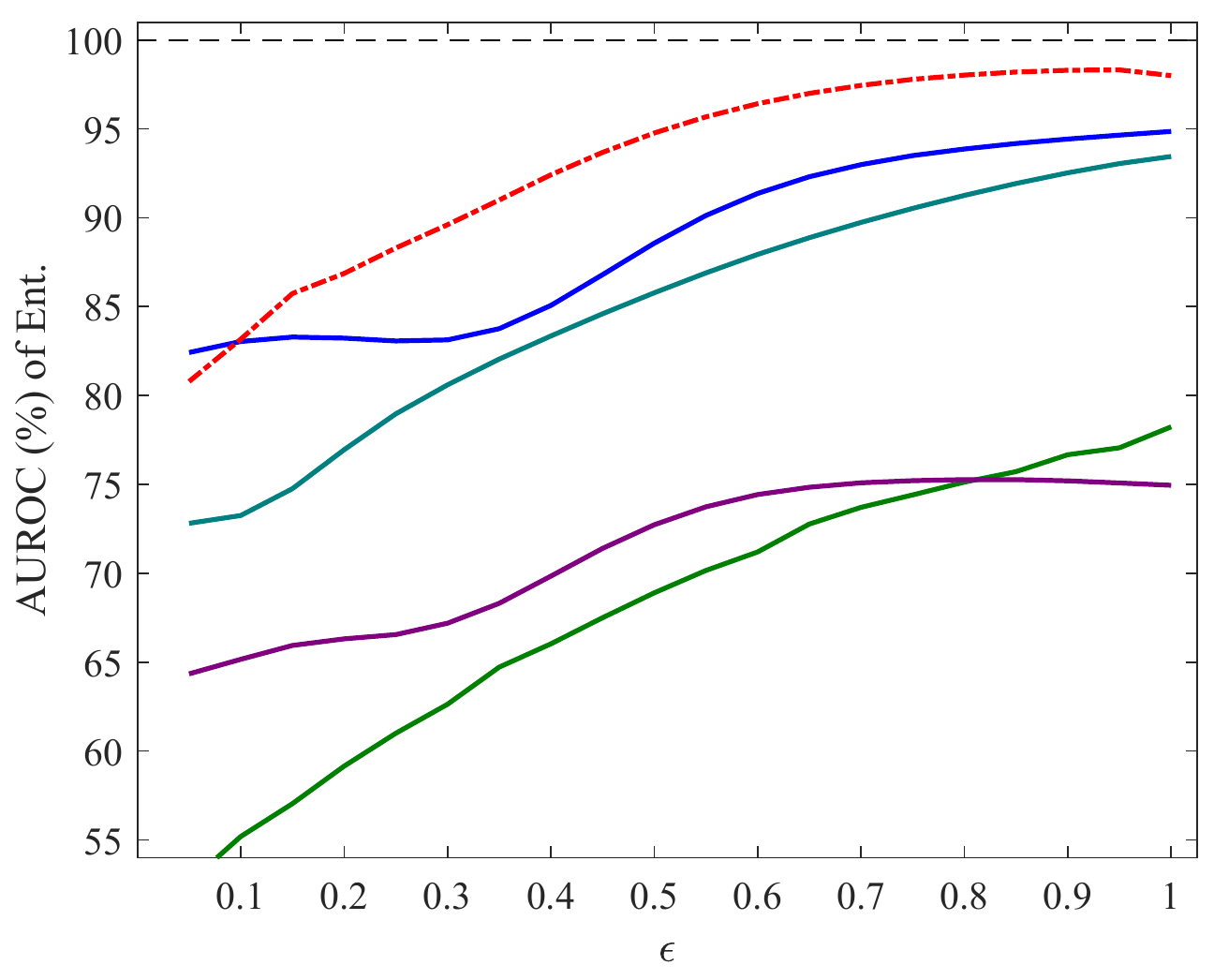}
        \subcaption{Ent. on CIFAR-$10$ }
    \end{subfigure}
    \begin{subfigure}[t]{0.49\linewidth}
        \centering
        \includegraphics[width=1\linewidth]{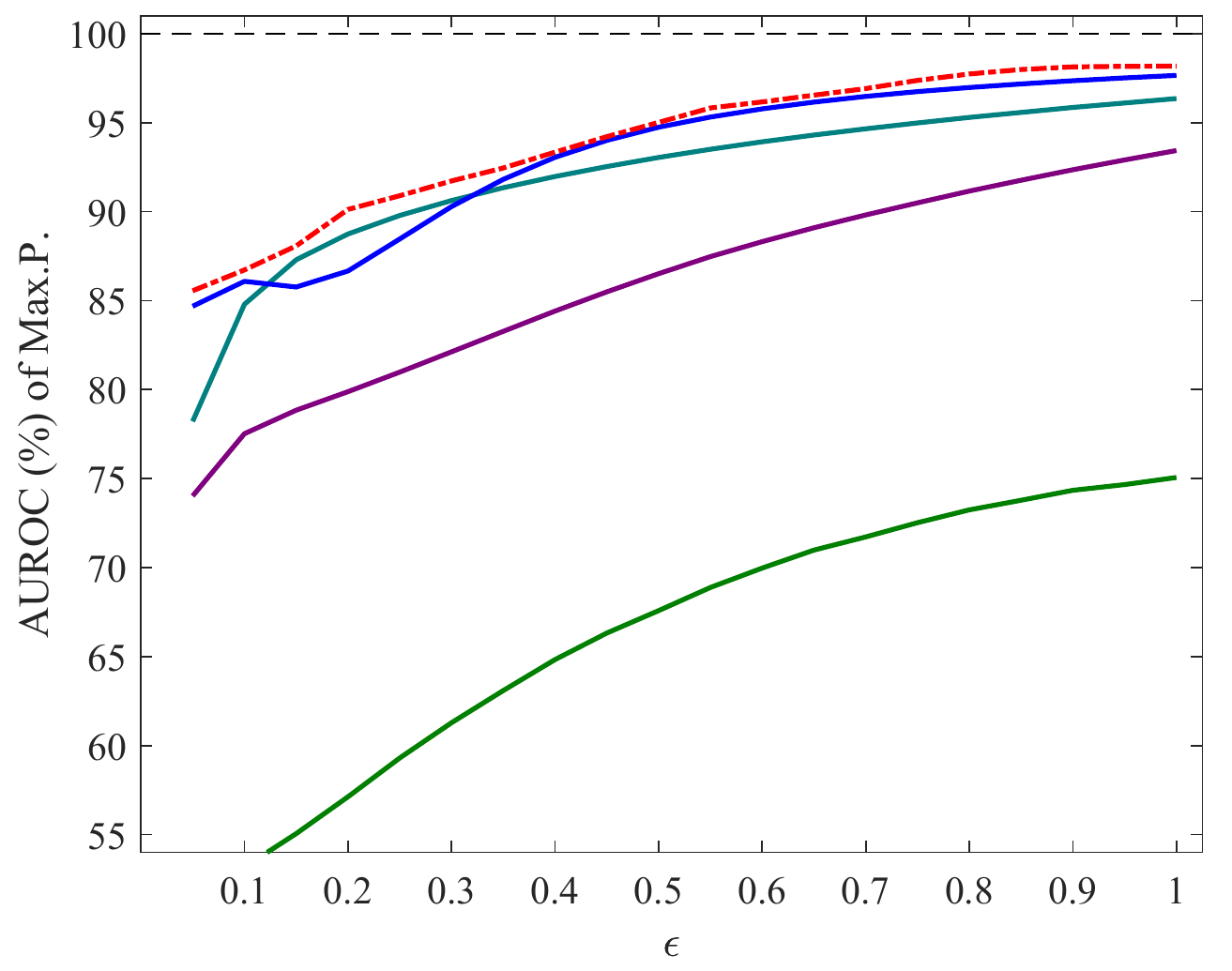}
        \subcaption{Max.P. on SVHN }
    \end{subfigure}
    \begin{subfigure}[t]{0.49\linewidth}
        \centering
        \includegraphics[width=1\linewidth]{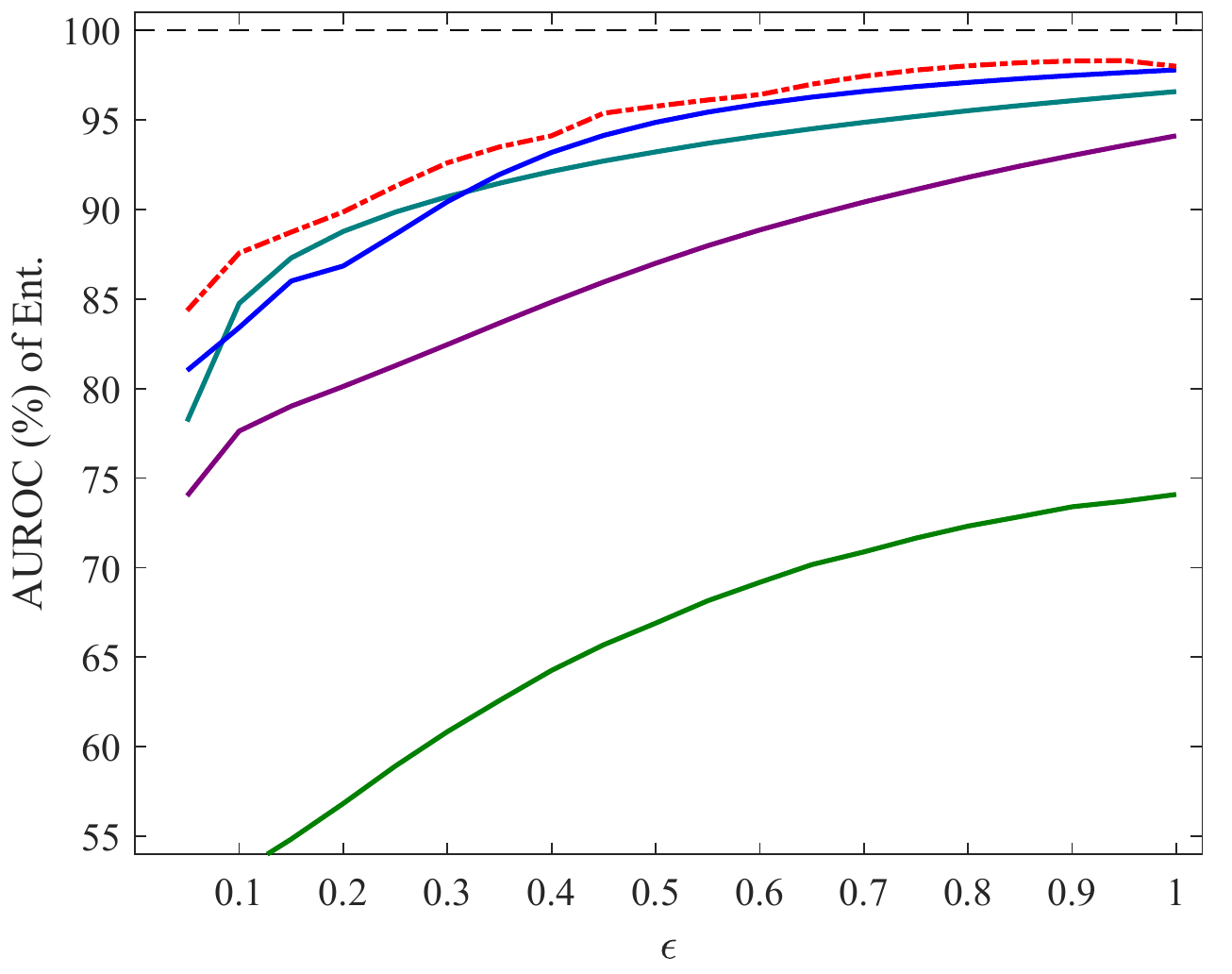}
        \subcaption{Ent. on SVHN }
    \end{subfigure}
    \end{tabular}}
    \caption{Performance on CIFAR-$10$ and SVHN datasets. $\epsilon$ is the step size in the FGSM and selected in the set $\{\frac{n}{20}\}_{n=1}^{20}$. The dashed and solid lines in each subfigure present the MU-UI and the referred methods, respectively.
    }\label{fig:cifar10}
\end{figure*}

\begin{figure}[!t]
    \centering
    \resizebox{\linewidth}{!}{
    \begin{tabular}{l}
    \includegraphics[width=\linewidth]{FIG/caption.pdf} \\
    \begin{subfigure}[t]{0.49\linewidth}
        \centering
        \includegraphics[width=1\linewidth]{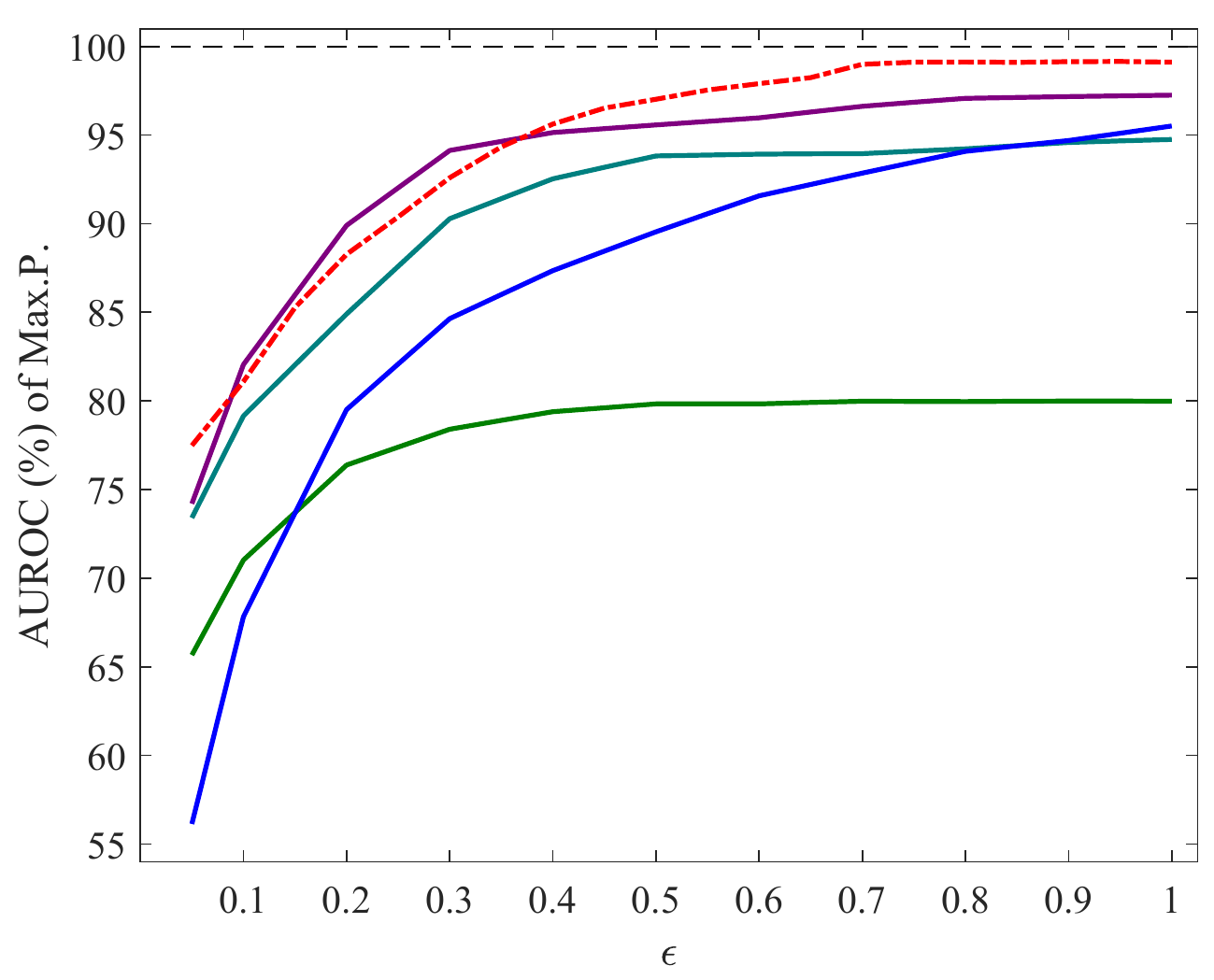}
        \subcaption{Max.P. with ResNet backbone}
    \end{subfigure}
    \begin{subfigure}[t]{0.49\linewidth}
        \centering
        \includegraphics[width=1\linewidth]{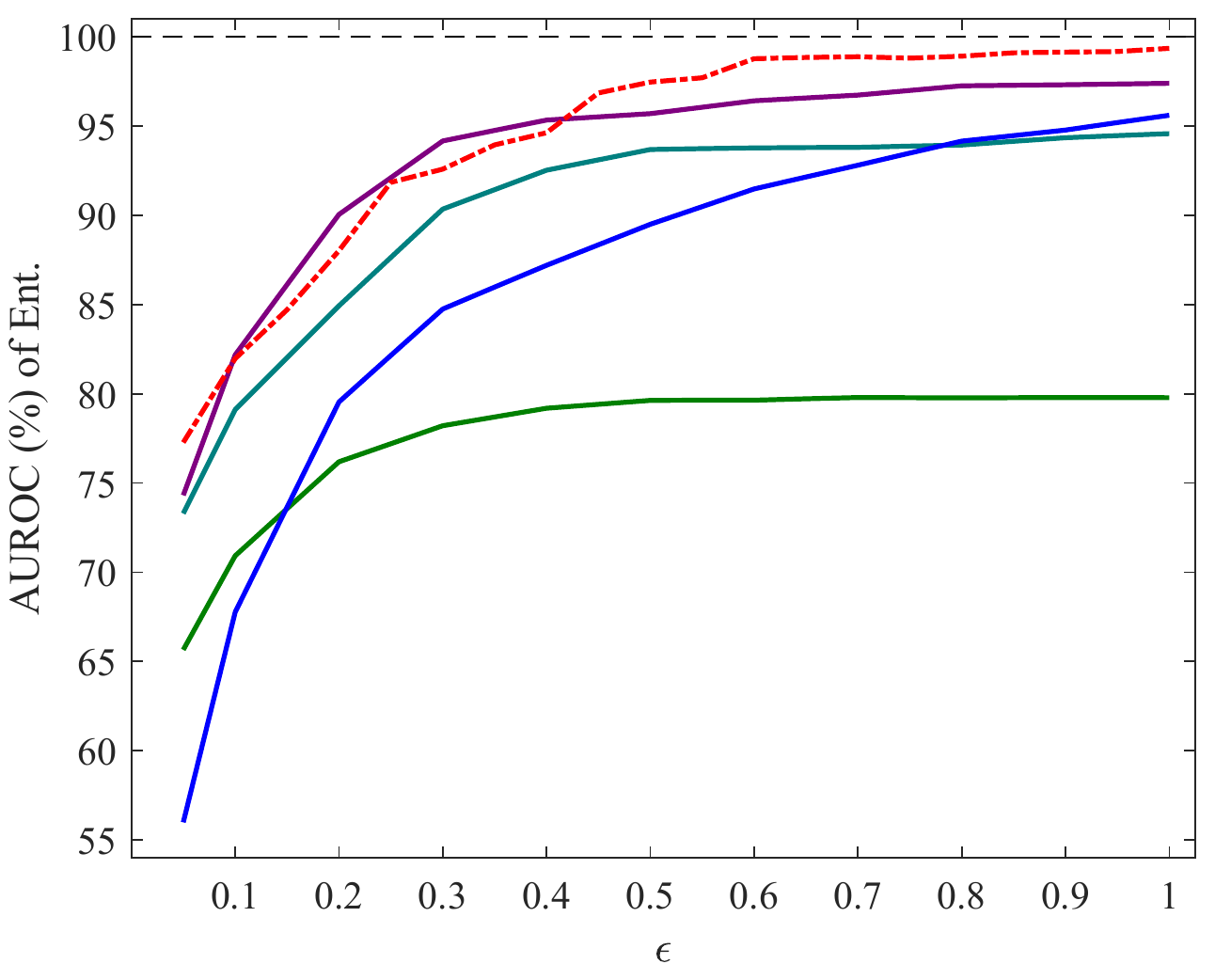}
        \subcaption{Ent. with ResNet backbone}
    \end{subfigure}
    \end{tabular}}
    \caption{Performance on ResNet$18$ backbone with CIFAR-$10$ dataset. $\epsilon$ is the step size in the FGSM and selected in the set $\{\frac{n}{20}\}_{n=1}^{20}$. $8$ heads were used to connect all basiclock in ResNet$18$. The dashed and solid lines in each subfigure present the MH-UI and the referred methods, respectively.
    }\label{fig:resnet}
    \vspace{-4mm}
\end{figure}

\section{EXPERIMENTAL RESULTS AND DISCUSSIONS}
\label{sec:experiments}
\subsection{Implementation Details}
\label{ssec:Implementation Details}

In our experiments, we used GoogLeNet~\cite{szegedy2014going} and ResNet~\cite{he2016deep} as the backbone and followed the experimental settings in~\cite{xie2021dsui,malinin2018predictive,malinin2019reverse}. We connected $9$ heads (excluding the final classifier) after each inception output for the GoogLeNet model and $7$ heads for the ResNet model. In the training procedure, we first trained the backbones. we applied Adam optimizer by following~\cite{malinin2018predictive} with $100$ epochs. We used $1$-cycle learning rate scheme, where we set initial learning rate as $7.5\times10^{-4}$ and cycle length as $70$ epochs for each dataset. All other referred methods were undertaken in the same training way. Then, we trained the heads in sequence. For the heads, the learning rate was fixed as $0.001$ with $50$ epoch for each dataset.
We compared the MH-UI with MC dropout~\cite{gal2016dropout}, DPN~\cite{malinin2018predictive}, RKL~\cite{malinin2019reverse}, and DS-UI~\cite{xie2021dsui} for adversarial attack detection in CIFAR-$10$~\cite{krizhevsky09cifar} and SVHN~\cite{netzer2011reading} datasets. 

We introduced max probability and entropy of the estimated Dirichlet distribution as metrics for uncertainty measurement and adopted the area under receiver operating characteristic curve (AUROC) ~\cite{malinin2018predictive,malinin2019reverse,xie2021dsui}. The larger the values of AUROC, the better the performance. For all the methods, we conducted five runs and reported the mean of AUROC.

\subsection{Ablation Study}
\label{ssec:Ablation Studies}

We conducted ablation study on the CIFAR-$10$ dataset under adversarial attack detection to discuss the head combination configurations, as well as the combination of head effect. We set five different combinations of heads in our experiment. Table~\ref{table3} shows the setting details and means of AUROC results. It can be observed that the model with lager number of heads used in the MH-UI performed better in most of instances. Thus, we choose the combination of all the head $\left\{1,2,3,4,5,6,7,8,9,10\right\}$ in the following experiment, which performed best in the ablation study.

\subsection{Adversarial Attack Detection}
\label{ssec:Attack Detection Results}

Table~\ref{table1} illustrates the accuracies on both CIFAR-$10$ and SVHN datasets of different heads in the proposed MH-UI and the referred methods. Different heads perform diversely and gradually better among the increasing depth, while the final accuracies of the last head in Table~\ref{table2} achieve comparable performance compared with the referred methods. The classification accuracies in Table~\ref{table2} is for convergence evaluation of model training in each method under clean image setting. And all the methods can converge well and obtain high accuracies. Our method can achieve competitive results and does not reduce the accuracy of the original framework. 

Figure~\ref{fig:cifar10}(a) and (b) illustrate the performance of each method in the adversarial attack detection for CIFAR-$10$ dataset. We used FGSM attack method in experiment and set $\epsilon$ in the FGSM in the set $\left\{\frac{n}{20}\right\}^{20}_{n=1}$. Comparing to different methods, the MH-UI converges to lager value  when $\epsilon$ grows larger ($\epsilon>0.2$). And as $\epsilon$ is small, the MH-UI can perform similar to the best one (RKL).

Figure~\ref{fig:cifar10}(c) and (d) illustrate experimental result of each method in the attacked sample detection on SVHN dataset. As the SVHN dataset needs to classify the digital features relatively simple, all methods perform better than those on CIFAR-$10$. The MH-UI work better in two evaluation criteria. And when $\boldsymbol{\epsilon}$ approaches to $1$, the AUROC of detecting attack can converge to $98\%$ in both Max.P. and Ent.

In summary, compared with other UI methods, the above results show that our MH-UI method is effective on detecting adversarial attack in different datasets.
 
\subsection{Generalized Applicability}
\label{ssec:general applicability}

Furthermore, our method can be applied to other models. We applied ResNet$18$~\cite{he2016deep} as backbone. Different from the head used in GoogLeNet, we used a three-layer FC net with $1024$ hidden units for each hidden layer in each one. For other comparing methods, we also built them with ResNet$18$ as backbone.  Figure~\ref{fig:resnet} illustrates each method's experimental result with ResNet backbone. Our method also works well in ResNet. When the samples are under heavy attack ($\boldsymbol{\epsilon}$ \textgreater$0.2$) the AUROC of Max.P. and Ent. can converge to $99\%$. These result shows that our method can also be used in other models backbone such as ResNet.

\section{CONCLUSION}

In this paper, we have proposed a multi-head uncertainty inference (MH-UI) framework for detecting adversarial attack examples. We adopted a multi-head architecture with multiple prediction heads to obtain predictions from different depths in the DNNs and introduced shallow information for the UI. Using independent heads at different depths, the normalized predictions are assumed to follow the same Dirichlet distribution and the distribution parameter of it is estimated by moment matching. The proposed MH-UI is flexible with different head numbers and backbone structures. Experimental results show that the proposed MH-UI framework can outperform all the referred UI methods in the adversarial attack detection task with different settings.

\bibliographystyle{IEEEbib}
\bibliography{strings}

\end{document}